\pdfoutput=1
\documentclass[12pt,a4paper]{article}
\usepackage[T1]{fontenc}
\usepackage{lmodern}
\usepackage[english]{babel}
\usepackage{graphicx}
\usepackage{caption}
\usepackage{amsmath}
\usepackage[toc]{appendix}
\usepackage{amsfonts}
\usepackage{natbib}
\usepackage{xcolor} 
\usepackage{booktabs}
\usepackage{colortbl}
\usepackage{array}
\usepackage{graphicx} 
\usepackage{makecell}
\usepackage{caption}
\usepackage[colorlinks, citecolor=blue]{hyperref}    
\usepackage{authblk} 
\usepackage{listings}
\usepackage{xcolor}
\usepackage{subcaption}

\title{Differential Contributions of Machine Learning and Statistical Analysis to Language and Cognitive Sciences}

\date{\vspace{8ex}} 
\author[1]{\footnotesize Kun Sun\thanks{\texttt{email: sharpksun@hotmail.com}}}
\author[2]{\footnotesize Rong Wang}

\affil[1]{\footnotesize Tübingen, Germany}
\affil[2]{\footnotesize The Institute of Natural Language Processing, Stuttgart University, Stuttgart, Germany}

\begin{document}	

\maketitle
\vspace{-2.6cm}

\begin{abstract}

	Data-driven approaches have revolutionized scientific research, with machine learning and statistical analysis being commonly used methodologies. Despite their widespread use, these approaches differ significantly in their techniques, objectives and implementations. Few studies have systematically applied both methods to identical datasets to highlight potential differences, particularly in language and cognitive sciences. This study employs the Buckeye Speech Corpus to illustrate how machine learning and statistical analysis are applied in data-driven research to obtain distinct insights on language production. We demonstrate the theoretical differences, implementation steps, and unique objectives of each approach through a comprehensive, tutorial-like comparison. Our analysis reveals that while machine learning excels at pattern recognition and prediction, statistical methods provide deeper insights into relationships between variables. The study highlights how semantic relevance, a novel metric measuring contextual influence on target words, contributes to understanding word duration in speech. We also systematically compare the differences between regression models used in machine learning and statistical analysis, particularly focusing on the training and fitting processes. Additionally, we clarify several common misconceptions that contribute to the confusion between these two approaches. Overall, by elucidating the complementary strengths of machine learning and statistics, this research enhances our understanding of diverse data-driven strategies in language and cognitive sciences, offering researchers valuable guidance on when and how to effectively apply these approaches in different research contexts.

\end{abstract}










\clearpage
\section{Introduction}
\label{introduction}
Advancements in technology have revolutionized the ability of scientists to gather and analyze data on an unprecedented scale. This surge in data availability has led to the necessity for more sophisticated quantitative techniques and computational tools, fundamentally changing the way research is conducted and facilitating new discoveries across various domains (\citealp{langley1981data}; \citealp{han1993data}; \citealp{kitchin2014big}; \citealp{montans2019data}; \citealp{jack2018data}; \citealp{nunez2019happened}). Data-based technologies serve as powerful illuminators, revealing hidden patterns and insights within vast datasets, much like a flashlight cutting through darkness. For example, machine learning (ML) has become indispensable in scientific inquiry, allowing researchers to decipher complex patterns, enhance their understanding of intricate phenomena, and make precise predictions. This technology is applied across diverse fields including medicine, astronomy, genomics, social sciences, and environmental studies, enabling researchers to swiftly extract significant insights from large volumes of data, much faster than traditional methods would permit (\citealp{mjolsness2001machine}; \citealp{jordan2015machine}; \citealp{lecun2015deep}; \citealp{webb2018deep}). 

Such acceleration not only deepens the exploration of research questions but also unveils knowledge that was once concealed or unreachable. On the other hand, statistical methods play a critical role in the modern landscape of scientific research. They are integral in every phase of a study, from planning and design to data collection, analysis, and the interpretation and reporting of results (\citealp{carleo2019machine}; \citealp{nunez2019happened}; \citealp{butler2018machine}; \citealp{grimmer2021machine}). While machine learning helps in identifying patterns that may elude human detection in massive datasets, statistical analysis continues to enrich scientific research by providing robust frameworks for making inferences and validating findings (\citealp{fisher1955statistical}; \citealp{box1976science}; \citealp{nelder1986statistics}; \citealp{nosek2012scientific}). Together, these tools are taken as powerful data-science tools, transforming scientific paradigms, propelling forward our capacity to understand and manipulate the natural world.

Data-driven research has made significant strides across various fields, bringing to the forefront disciplines such as data science, statistics, machine learning, and deep learning (\citealp{solomatine2008data}; \citealp{wolf2010data};  \citealp{miller2015data}; \citealp{zhang2011data}). Among these, machine learning and statistics form the foundational core in data science. While often conflated due to their shared capability to process data, machine learning and statistics frequently overlap, creating ambiguity around their distinct roles and unique contributions. Moreover, both machine learning and statistics take advantage of the shared knowledge of probability theory, linear algebra etc. Contemporary research methodologies frequently incorporate machine learning and statistical analysis, often employing both to harness the strengths of each approach. Many researchers have found it challenging to distinguish clearly between these two areas. Although both fields are fundamentally concerned with extracting knowledge from data, they differ significantly in their objectives, approaches and implementations. Gaining a clear understanding of these differences is crucial for mastering both fields and effectively applying appropriate data-driven strategies. 


Despite the prevalent combined use of the two methods, there remains a significant gap in studies that systematically apply both techniques to identical datasets to highlight potential differences in outcomes. The systematic comparisons between them have been rarely taken in social sciences and behavior research, and such comparisons could provide valuable insights into how each method processes and interprets the same dataset differently. Moreover, the specific contributions of these techniques to these fields have not been extensively explored. In particular, the fields of language sciences, cognitive research, and social sciences could greatly benefit from targeted studies examining how machine learning and statistical analysis can uniquely advance our understanding of complex phenomena within these disciplines. Such investigations could elucidate the distinct advantages or limitations of each method, potentially leading to more refined and effective data-driven methodologies in these areas. 

The present study aims to bridge existing gaps in research methodology by applying both statistical analysis and ML techniques to the same dataset. By demonstrating theoretical differences, distinct implementation steps and achieving different objectives, we provide a comprehensive, tutorial-like comparison of these approaches. This detailed examination offers researchers in cognitive and social sciences not only theoretical insights into the differences between machine learning and statistics but also hands-on experience in implementing these methods. 
The current study illustrates the unique applications of each method in addressing relevant questions, implementing comprehensive procedures, and extracting valuable insights. By doing so, we aim to enhance the understanding of when and how to effectively apply these data-driven approaches in various research contexts for both researchers and practitioners. This study not only elucidates the theoretical differences between the two data-driven approaches but also provides effective practical implementations for each. It will be particularly valuable for empirical and data-driven research in the social and cognitive sciences.


\section{Background: Relation and differences between statistics and machine learning}
Before delving into our demonstration, it is crucial to systematically examine the relationship between statistics and ML, highlighting both their connections and distinctions from a theoretical perspective. By thoroughly understanding these differences, we believe that the practical implementations will be significantly enhanced. This foundational knowledge will provide a solid framework for applying these concepts effectively in real-world scenario.

\subsection{How are statistics and machine learning related?}


The early 20th century saw statistics rapidly advance, empowering scientists to quantitatively assess hypotheses like ``Does treatment X affect outcome Y?''. This led to the development of tools focused on hypothesis testing and confidence intervals, aimed at providing clear answers and estimating effect sizes. In contrast, machine learning emerged from engineering with the goal of enhancing machine capabilities. Initially, it focused on improving machine functionality and, secondarily, on understanding intelligence itself. Unlike statistics, machine learning was not primarily concerned with verifying real-world truths; instead, these were seen as potential components of intelligent behaviors--a perspective still debated today. This fundamental difference in origins and objectives highlights the distinct approaches of statistics and machine learning, despite their overlapping applications in data analysis and prediction.

\citet{breiman2001statistical} highlighted the divergent approaches to data analysis, emphasizing the ongoing evolution of these methodologies over the decades. Many machine learning techniques have their roots in traditional statistical methods, such as linear and logistic regression, while also drawing from disciplines like calculus, linear algebra, and computer science. This intersection has led some to mistakenly merge the concepts of machine learning and statistics. Even though they started out separately, machine learning often uses statistical ideas that have been around for over 100 years. These math principles work the same whether you're trying to invent artificial intelligence, do research, or create fair ways to measure things. Because of this, many big questions in machine learning are actually old statistical problems that have not been looked at much before.

Moreover, the introduction of user-friendly machine learning packages, such as Python's \texttt{scikit-learn} \citep{pedregosa2011scikit}, has further abstracted machine learning from its statistical foundation. This has propagated a belief among some newcomers to the field that a deep understanding of statistics is not necessary for machine learning applications. While basic tasks might not require intensive statistical knowledge, advanced modeling and the development of new algorithms heavily rely on a solid grounding in statistics and probability theory.



\subsection{How are two fields different?}


 In recent years, statistics has been both pushed and energized by machine learning's success, especially in areas like predicting outcomes, which used to be mainly statistics' territory (\citealp{l2017machine}; \citealp{saidulu2017machine}; \citealp{ratner2017statistical}; \citealp{rudin2022interpretable}; \citealp{ratner2017statistical}; \citealp{ley2022machine}). This has sparked vigorous efforts to merge the theories and tools of both fields. However, for these efforts to be successful, they must first recognize and address the underlying reasons for their differences (\citealp{bzdok2018statistics}; \citealp{makridakis2018statistical}; \citealp{boulesteix2014machine}). Understanding these differences is essential for employing the strengths of both fields effectively, whether the goal is to discover deep insights from data or to develop robust predictive algorithms.

One of the fundamental distinctions between statistics and machine learning is their \textbf{purpose}. Statistics aims to infer properties about a population through samples, focusing on understanding and describing the relationships among variables in data. In contrast, machine learning seeks to predict outcomes based on patterns identified in data, often using large and complex datasets to train predictive models. This training process typically involves dividing data into subsets for training, validation, and testing, which helps refine the models for better accuracy.

Another significant difference lies in how data is \textbf{approached}. In statistics, the focus is on the quality of data and the validity of the conclusions drawn from it through significance testing. Advanced statistical models need to learn from the data by fitting. In contrast, machine learning, however, emphasizes the quantity of data, often requiring large datasets to train models to achieve the accuracy needed for effective predictions. Despite this, the fitting and training processes differ in their objectives, strategies, and evaluation criteria (More details can be seen in Discussion Section 4.2).

They employ distinct \textbf{evaluation standards and validation methods}. Statistical regression commonly uses metrics like \textit{p}-values, R-squared, AIC (Akaike Information Criterion), and confidence intervals. In contrast, machine learning often uses metrics like accuracy, precision, recall, F1-score for classification, or RMSE (Root Mean Square Error) for regression tasks. Statistical analysis often relies on assumptions and theoretical properties of the model. However, machine learning emphasizes empirical validation, using techniques like cross-validation and hold-out test sets.

\textbf{Interpretability} also varies greatly between the two. Statistical models, often simpler and based on fewer variables, tend to be more interpretable. This clarity comes from the use of statistical significance tests that validate the relationships within the data. Machine learning models, in contrast, can become highly complex, especially with the inclusion of many variables, making them accurate yet sometimes difficult to decipher. This complexity can render machine learning models as ``black boxes'' (\citealp{adadi2018peeking}; \citealp{gilpin2018explaining}; \citealp{linardatos2020explainable}), where it is challenging to trace how inputs are transformed into outputs.

\begin{figure}
	\vskip 0.2in
	\begin{center}
		\centerline{\includegraphics[width=0.86\textwidth]{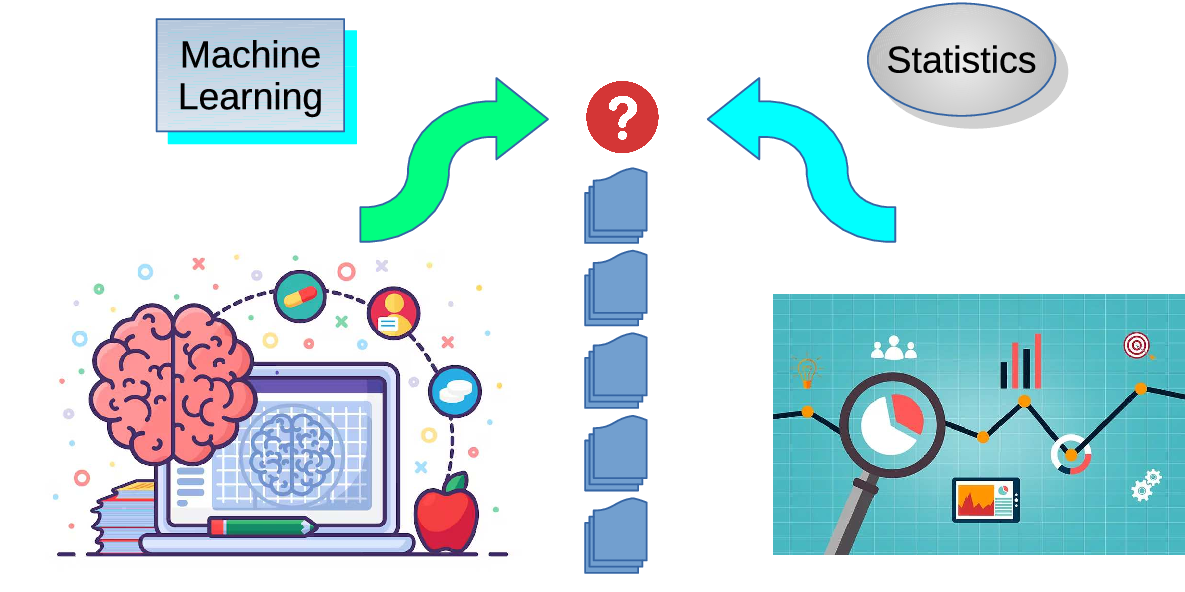}}
		\caption{Differences between machine learning and statistics. Note: Fig.~\ref{diff} could be aligned with Table~\ref{tab:ml_stats_differences}, Table~\ref{table:ml_steps}, and Table~\ref{table:gamm_steps}, making it easier to understand it.}
		\label{diff}
	\end{center}
	\vskip -0.26in
\end{figure}

In short, the contrast between machine learning and statistics is rooted in their different \textbf{objectives}, \textbf{approaches}, \textbf{evaluation standards}, and \textbf{priorities}. The key differences are summarized as shown in (a) of Table \ref{tab:ml_stats_differences} and Fig.\ref{diff}.

The following provides further key differences on technique characteristics. Machine learning is primarily focused on achieving high predictive accuracy and is comfortable with models that are effective but not easily interpretable, often referred to as ``black box'' models. On the other hand, statistics places a greater emphasis on relation among variables in data. Statisticians value models that are transparent and explainable, reflecting the field's deep roots in mathematics and science, where theoretical foundations and provable properties are paramount. This includes a strong focus on the behavior of estimates as sample sizes increase and ensuring that models are robust even with small datasets. This is summarized in (b) of Table \ref{tab:ml_stats_differences}. 


\begin{table}[ht]
	\centering
	\caption{Key Differences between ML and Statistics}
	\label{tab:ml_stats_differences}
	
	\begin{subtable}{\textwidth}
		\centering
		\caption{Differences on Implementing ML and Statistics}
		\scalebox{0.8}{
			\begin{tabular}{@{}lll@{}}
				\toprule
				\textbf{Procedure} & \textbf{Machine Learning} & \textbf{Statistics} \\ \midrule
				\textbf{Purpose} & Patterns for classification, clustering and regression & Relationships among variables \\
				\textbf{Approach} & Train ML models & Fit in statistical models \\
				\textbf{Evaluation} & Precision, recall, F1-score & \textit{p}-value, AIC \\
				\textbf{Outcome} & ML model to automatically analyze new data & Clarify the relationship among variables \\
				\bottomrule
			\end{tabular}
		}
		
		\label{subtab:implementation}
	\end{subtable}
	
	\vspace{0.3cm} 
	
	\begin{subtable}{\textwidth}
		\centering
		\caption{Technical Differences}
		\scalebox{0.8}{
			\begin{tabular}{@{}lll@{}}
				\toprule
				\textbf{Characteristic} & \textbf{Machine Learning} & \textbf{Statistics} \\ \midrule
				\textbf{Predictive accuracy} & High & Lower \\
				\textbf{Black box models} & High & Low \\
				\textbf{Relation among variables} & Low & High \\
				\textbf{Various asymptotic properties} & Low & High \\
				\textbf{Provable characteristics and bounds} & Low & High \\
				\bottomrule
			\end{tabular}
		}
		
		\label{subtab:technical}
	\end{subtable}
\end{table}

\subsection{Exemplification: Children's cognitive/language development}
\label{exem}
Linear regression is fundamentally a statistical method designed to minimize the squared error between data points. However, linear regression is also widely used in machine learning for predictive tasks. An example of employing regression models in machine learning or statistical analysis is taken to illustrate different intentions and outcomes in language sciences. 

In language sciences, regression models can be particularly illuminative. For example, researchers might use regression analysis to explore how the complexity of syntactic structures in children's language development (or other linguistic factors, such as vocabulary diversity, sentence length, speech rate or fluency measures) correlates with cognitive development indicators (indicated by scores on cognitive tests) (\citealp{goldin2014new}; \citealp{unal2020relations}; \citealp{monshizadeh2021there}). Statistically, this would involve collecting data on children's language use and cognitive tests, then applying linear regression to understand and quantify how changes in complexity of syntactic structure relate to children cognitive development, typically across a complete dataset. The statistical approach focuses on inference—determining whether there is a statistically significant relationship between syntactic complexity and cognitive development. This method does not necessitate training and testing subsets of data; instead, it aims to characterize the relationship across the entire data set, assessing the significance and reliability of the observed relationships.

Conversely, in a machine learning context, the same regression model could be employed to predict cognitive developmental outcomes based on the existing linguistic data (e.g., syntactic complexity, age, vocabulary diversity, sentence length, speech rate or fluency measures etc.) \citep{briglia2021logistic}. The model would be trained on a subset of data, with the model's parameters adjusted to optimize performance as measured on a separate test set. The objective is less about understanding the underlying dynamics of the relationship and more about achieving high predictive accuracy.

Suppose a language scientist wants to predict future cognitive development based on early linguistic behaviors using machine learning. In that case, they might employ a regression model on a divided dataset (training and testing), focusing on how well the model predicts developmental outcomes in new, unseen data. This approach would likely prioritize predictive power over interpretability, with model adjustments driven by performance metrics on the test set. Thus, while both fields use regression models, the context and objectives dictate their implementation: statistical analysis seeks to uncover and explain relationships within the data, providing comprehensive insights into language development laws. In contrast, machine learning employs regression models primarily for their predictive capabilities, often in applications that require rapid assessments of developmental trends based on linguistic inputs.


These distinctions highlight the importance of differentiating between the aims and approaches of statistics and machine learning in exploring linguistic and cognitive questions. It is crucial to ensure that the chosen approach aligns with the specific goals of the research or application. In the following Experiment section, we will detail the implementation procedures, objectives, and outcomes of both machine learning and statistical analysis when applied to the same dataset. This comparison will provide a clearer understanding of how regression models are used differently in this example: studying children's language development. In the Discussion section, we will further explore the distinctions between regression models employed in machine learning and statistical analysis, highlighting their unique characteristics and applications.

\section{Experiments}
Having systematically explored the theoretical differences between machine learning and statistics, we conducted experiments to demonstrate their distinct implementations and unique contributions to linguistic and cognitive research. These practical demonstrations aim to illustrate how each approach is implemented and the specific insights they offer in these fields.

\subsection{Materials \& methods}
The \textbf{Buckeye Corpus} of conversational speech contains high-quality recordings from 40 speakers in Columbus OH conversing freely with an interviewer \citep{pitt2005buckeye}. The speech has been orthographically transcribed and phonetically labeled.  The sessions were conducted as sociolinguistics interviews, and are essentially monologues. The speech has been orthographically transcribed and phonetically labeled. The corpus includes 357,908 words. 

The original dataset from the Buckeye corpus provided diverse factors: speaker’s gender (female, male), age (old, young),   word duration, etc. However, we can add more factors such as  PoS tag for each word, word length, word frequency, the phrase rate, deletions and semantic relevance. Most of these factors have been investigated to show that they are closely related with word duration (\citealp{cowan1997there}; \citealp{baker2009variability}; \citealp{terroir2014perceptual}; \citealp{cohen2015informativity}; \citealp{pierrehumbert2016phonological}; \citealp{burki2018variation}). However, these relevant research mostly employed statistical correlation and simply linear regression to explore how these factors affect word duration in speech. In this study, we employ both machine learning techniques and advanced mixed-effect regression models to analyze the data in the Buckeye corpus, aiming to gain complementary insights from these distinct approaches. The specific information on these factors is detailed in the Table \ref{tab2}:

\begin{table}[ht]
	\centering
	\caption{Factors and their descriptions in the the Buckeye corpus }
	\begin{tabular}{@{}>{\raggedright\arraybackslash}p{4cm}>{\raggedright\arraybackslash}p{8cm}@{}}
		\toprule
		\textbf{Factor/Feature} & \textbf{Description} \\ 
		\midrule
		Word Duration & The time to articulate a target word in spontaneous speech. \\
		\hline
		Word Length & The number of alphabets in a target word. \\
		\hline
		(log) Word Frequency & (Logarithm of) the normalized frequency of the word in the subtitle corpus \footnote{\url{https://invokeit.wordpress.com/frequency-word-lists/}}. \\
		\hline
		CiteLength (i.e., Syllable Number) & The number of syllables in transcript phonetic form of this word. \\
		\hline
		PhraseRate (related to Speaking Rate) & [Word number in this phrase] / [duration from the beginning of the phrase to the end of the phrase in a target word]. \\
		\hline
		Deletion (related to Phonological Reduction) & The number of segments in a target word deleted or reduced. \\
		\hline
		Semantic Relevance & The semantic relatedness degree between the target word and its context (see the \textbf{Appendix A}). \\
		\hline
		Speaker/Sex/Age & Speakers in corpus; Female vs. male; young vs. old. \\
		\bottomrule
	\end{tabular}
	\label{tab2}
\end{table}

Additionally, semantic relevance is a novel metrics, representing how a target word is semantically related with the context. It measures the semantic degree of how the contextual information influence the target word (\citealp{sun2023interpretable}; \citealp{sun2023optimizing}). This metrics we introduced has proven effective in predicting eye-movements during reading multiple languages \citep{sun2023optimizing}, which encompasses a facet of language comprehension. This metrics is also capable of predicting and elucidating neural activity associated with the processing of naturalistic discourse, such as electroencephalography (EEG) \citep{sun2024eeg}. The computation of semantic relevance is seen in the \textbf{Appendix A}. 

\subsection{Study 1: Machine learning methods}

ML can be classified in various ways based on different criteria. One common categorization divides ML into three main types: supervised learning, where models predict an outcome based on input data; unsupervised learning, where models identify patterns and relationships in data without any specific outcome to predict; and reinforcement learning, where an agent learns to make decisions by receiving rewards for actions (\citealp{bishop2006pattern}; \citealp{hastie2009elements}; \citealp{murphy2012machine}; \citealp{harrington2012machine}; \citealp{raschka2019python}). The general mathematical equation for machine learning could be formulated as follows:

\begin{equation}
	y = f(\mathbf{X}) + \varepsilon 
\end{equation}

Where $y$ is the target variable (output),  $\mathbf{X}$ is the input variable or feature vector $\mathbf{X} = [x_1, x_2, \ldots, x_n]$,  $f$ is the function we are trying to learn or approximate, $\varepsilon$ represents the error term or noise. Basically, the goal of machine learning is to optimize the parameters in Equation (1). The key to optimize the parameters is to reduce the error or loss. From this perspective, the main purpose of training a machine learning model is to minimize its loss function. The loss function $L$ can be expressed as:

\begin{equation}
	L = \sum_{i=1}^{n} (y_i - f(\mathbf{X}_i))^2
\end{equation}

ML can accomplish three primary tasks: clustering, classification, and regression. Various typical machine learning models exist, including linear regression, decision trees, random forests, support vector machines (SVMs), K-nearest neighbors (KNN), Naive Bayes, K-Means, and principal component analysis (PCA). Each model has its unique strengths that make it particularly suited for specific tasks. For instance, K-Means excels at clustering, while random forests and decision trees are adept at classification. Linear regression, on the other hand, is particularly effective for regression tasks.

When discussing specific models, their functions can indeed be more specialized compared to their general purpose. For linear regression exampled in the section~\ref{exem}, $f(\mathbf{X})$ might take the form:

\begin{equation}
	f(\mathbf{X}) = w_0 + w_1x_1 + w_2x_2 + \ldots + w_nx_n
\end{equation}






After grasping the fundamental principles of machine learning, we can outline the basic steps for implementing machine learning to perform classification or regression tasks. These steps can be summarized in Table~\ref{table:ml_steps}.

\begin{table}[!ht]
	\centering
	\caption{Steps for implementing ML models}
	\begin{tabular}{|c|l|p{6cm}|}
		\hline
		\textbf{Step} & \textbf{Implementation} & \textbf{Description} \\
		\hline\hline
		1 & Define problem and gather Data & Clearly state the problem(s) you're trying to solve and collect relevant, high-quality data. \\
		\hline
		2 & Preprocess data and select features & Clean the data, handle missing values, and understand patterns and relationships. Select useful features from the data.  \\
		\hline
		3 & Choose model & Select an appropriate machine learning algorithm based on the problem type (e.g., classification, clustering, regression). \\
		\hline
		4 & Split data & Divide your dataset into training and testing sets. \\
		\hline
		5 & Train model & Use the training data to teach the chosen algorithms to make predictions or decisions. \\
		\hline
		6 & Evaluate model & Test the model's performance using the testing data and appropriate metrics (e.g., Precision, Recall, F1-score, Mean Absolute Error etc. )\\	\hline
		7 & Tune and optimize & Adjust (hyper)parameters and refine the model using some strategies (e.g., bootstrap, normalizing the data etc.) to improve its performance. \\
		\hline
		8 & Deploy and Monitor (optional) & Integrate the model into the application or system, and continuously monitor its performance in real-world conditions. \\
		\hline
	\end{tabular}
	
	\label{table:ml_steps}
\end{table}

Following the steps outlined in Table~\ref{table:ml_steps}, we can detail the implementation of a machine learning task using the \texttt{Buckeye corpus}. Initially, we define the problem. Machine learning will be employed for clustering, classification, or regression tasks. Given the characteristics of the Buckeye corpus, our goal is to employ supervised machine learning models to identify patterns in new data. The primary objective of this ML task is to predict the ranges of word speech duration, which is a typical \textbf{classification} problem. To alleviate the training burden, we categorize word durations into eight distinct ranges. Originally, there are 357,908 words, each with a unique duration. However, we have grouped these 357,908 unique values into eight categories. Essentially, using other features in the dataset, the chosen ML models are intended to predict which of the eight categories a word's duration falls into.

Second, referring to Step 3 in Table~\ref{table:ml_steps}, we could introduce several machine learning models to accomplish the classification task. A variety of models could be selected to implement this classification. Based on the nature of the task and the dataset, we have chosen the Random Forest (RF) and support vector machine (SVM) models to perform the classification. The reason for this is that both RF and SVM perform well in classification tasks, making them typical and reliable ML models for this purpose. Another reason is the specific characteristics of the Buckeye Corpus dataset.

Third, to train an effective ML model, we must carefully select relevant features from the dataset. For this study, we believe that the factors from the Buckeye corpus such as age, gender, word length, word frequency, and speaking rate, among others, could be particularly valuable. Once we have identified these key features, we'll divide our dataset into two portions: 75\% for training and 25\% for testing. The larger training set will be used to train our ML model, while the smaller testing set will be reserved for evaluating the model's performance on unseen data. This approach ensures an unbiased assessment of the model's effectiveness and generalizability.


Lastly, in most cases, we need to implement strategies to optimize the model to enhance the evaluation results. Probably we could decrease or increase some given features in the model to improve the model's performance. In most cases, some strategies for each speficic models could be considered. This step can help minimize the loss function described in Equation (2), thereby optimizing the parameters in Equation (1) for improved outcomes.

The following sections demonstrate the implementation of RF and SVM models, illustrating how to implement these steps in practice. We also provide the corresponding Python scripts to execute these steps.

\subsubsection{Random forest}

Random Forest is a sophisticated ensemble machine learning algorithm that constructs multiple decision trees during the training phase, addressing the overfitting tendencies often associated with individual decision trees \citep{breiman2001random}. This method operates by training each tree on a random subset of the data and selecting a random subset of features at each node to determine the optimal split. For predictions, Random Forest employs a democratic approach: in classification tasks, each tree votes, and the majority determines the final class assignment, while in regression tasks, the algorithm calculates the mean of individual tree predictions. This aggregation methodology serves to mitigate errors from individual trees and leverage the collective strength of multiple learners, resulting in enhanced robustness against overfitting. The algorithm's performance is typically evaluated using standard metrics such as accuracy for classification tasks. 
This versatile approach has found wide application across various domains, including finance, healthcare, and environmental sciences, due to its ease of use, ability to handle both categorical and continuous variables, built-in feature importance ranking, and efficiency in processing large datasets.

The subsequent Python script demonstrates the implementation of a classification task for different duration classes. This code exemplifies the specific steps required to develop and deploy a machine learning model for the classification purpose designed for predicting the ranges of word speech duration in the \texttt{Buckeye corpus}, as outlined in Table~\ref{table:ml_steps}. The script serves as a practical illustration of the theoretical framework discussed earlier, showcasing how each step in the machine learning process is translated into executable code.

\begin{lstlisting}[caption={Python code for loading data and training a RandomForest classifier}, label=lst:randomforest]
	import pandas as pd
	import numpy as np
	from sklearn.model_selection import train_test_split
	from sklearn.ensemble import RandomForestClassifier
	from sklearn.metrics import accuracy_score
	import matplotlib.pyplot as plt
	from sklearn import tree
	
	# Load the data
	df = pd.read_csv("buckeye1.csv", delimiter='\t')
	
	# Define the breaks for the ranges of 'Duration'
	breaks = np.quantile(df['Duration'], np.arange(0, 1, 0.2))
	
	# Define labels
	labels = ['1', '2', '3', '4', '5']  # Number of labels matches number of quantile edges minus one
	
	# Cut the 'Duration' variable into ranges and assign labels
	df['range_label'] = pd.cut(df['Duration'], bins=breaks, labels=labels, include_lowest=True)
	
	# Drop rows with missing values in 'range_label'
	df.dropna(subset=['range_label'], inplace=True)
	
	# Feature selection
	feature_names = ['CiteLength', 'PhraseLength', 'Deletions', 'WordLength', 'LogWordFreq']
	X = df[feature_names]
	y = df['range_label']
	
	# Split the data into training and testing sets
	X_train, X_test, y_train, y_test = train_test_split(X, y, test_size=0.25, random_state=42)
	
	# Create and train the Random Forest Classifier
	clf = RandomForestClassifier(n_estimators=100, random_state=42)
	clf.fit(X_train, y_train)
	
	# Make predictions
	preds_rf = clf.predict(X_test)
	accuracy = accuracy_score(y_test, preds_rf)
	print("Accuracy:", accuracy)
	# Accuracy: 51.02712%
\end{lstlisting}

The RF model using Python's \texttt{matplotlib.pyplot} and \texttt{sklearn.tree} libraries. This kind of visualization can be particularly useful for understanding how individual trees in the ensemble make decisions, which can provide insights into the model's operation. 

However, we can streamline the factors (features) in the training dataset. For instance, by selecting only `WordLength' and `WordFrequency', the prediction accuracy achieved was 50.75\%. Conversely, using `CiteLength' and `PhraseRate' as factors, the prediction accuracy dropped to 37.13\%. Furthermore, combining `Deletions' with `WordLength' yielded a prediction accuracy of 42.36\%. By experimenting with various factor combinations, we can deduce the impact of different factors on machine learning accuracy. Evidently, `WordLength' and `WordFrequency' appear to play more significant roles in enhancing machine learning performance. 

To enhance the performance of our ML model, we explored the impact of incorporating additional features. We expanded our feature set to include `Age': feature\_names = [`Age', `CiteLength', `PhraseLength', `Deletions', `WordLength', `LogWordFreq'].  Following the same training procedure with this expanded feature set, we observed an improvement in the accuracy score from 51\% to 53.34\%. Further, when we included `Speaker' as an additional feature, the accuracy score increased to 59.27\%. These results show the critical role that feature selection plays in optimizing ML model performance. By carefully choosing and incorporating relevant features, we can significantly enhance the performance of the trained ML models. Specifically, the factor `Speaker' is a signficant feature in training ML classifier. 



Despite of some progress in classification, we still need to introduce some optimized strategies to improve the model performance, that is, we could improve the accuracy score, and this step is ``Tune and optimize'' in  Table~\ref{table:ml_steps}. These strategies for optimizing the RF model include the following. First, feature scaling is introduced to normalize the features and it may improve the model's accuracy. Second, the implementation of hyperparameter tuning through the function of ``GridSearchCV'' allows for the optimization of various parameters in this Random Forest Classifier, ensuring that the model is fine-tuned for better predictive performance. Third, we employ feature selection with the function of ``SelectFromModel'', identifying and retaining only the most important features based on their importance scores, which can lead to a more efficient and effective model.

Furthermore, the evaluation of model performance could be more comprehensive. To this end, we include a classification report that details precision, recall, and F1-score for each class, providing deeper insights into how well the model performs across different categories. To aid in understanding feature significance, a bar plot visualizing feature importances is also included. Lastly, the breaks calculation for creating duration ranges has been slightly adjusted to ensure that the upper bound is included. Overall, these changes reflect best practices in machine learning and are likely to yield improved accuracy and interpretability in the model's results. The implementation script is seen below.

\begin{lstlisting}*[caption={Python code for optimizing the RandomForest classifier}, label=lst:randomforest1]
	import pandas as pd
	import numpy as np
	from sklearn.model_selection import train_test_split, GridSearchCV
	from sklearn.ensemble import RandomForestClassifier
	from sklearn.preprocessing import StandardScaler
	from sklearn.metrics import accuracy_score, classification_report
	import matplotlib.pyplot as plt
	from sklearn.feature_selection import SelectFromModel
	
    ##the same code
    
	# Scale the features
	scaler = StandardScaler()
	X_train_scaled = scaler.fit_transform(X_train)
	X_test_scaled = scaler.transform(X_test)
	
	# Create and train the Random Forest Classifier with hyperparameter tuning
	param_grid = {
		'n_estimators': [100, 200, 300],
		'max_depth': [None, 10, 20, 30],
		'min_samples_split': [2, 5, 10],
		'min_samples_leaf': [1, 2, 4]
	}
	
	rf = RandomForestClassifier(random_state=42)
	grid_search = GridSearchCV(estimator=rf, param_grid=param_grid, cv=5, n_jobs=-1, verbose=2)
	grid_search.fit(X_train_scaled, y_train)
	
	# Get the best model
	best_rf = grid_search.best_estimator_
	
	# Feature importance-based selection
	selector = SelectFromModel(best_rf, prefit=True)
	X_train_selected = selector.transform(X_train_scaled)
	X_test_selected = selector.transform(X_test_scaled)
	
	# Train the model with selected features
	best_rf.fit(X_train_selected, y_train)
	
	# Make predictions
	preds_rf = best_rf.predict(X_test_selected)
	accuracy = accuracy_score(y_test, preds_rf)
	print("Accuracy:", accuracy)
	
	# Print classification report
	print("\nClassification Report:")
	print(classification_report(y_test, preds_rf))
	
	# Plot feature importances
	importances = best_rf.feature_importances_
	indices = np.argsort(importances)[::-1]
	selected_features = selector.get_support()
	
	plt.figure(figsize=(10,6))
	plt.title("Feature Importances")
	plt.bar(range(X_train_selected.shape[1]), importances[indices][:X_train_selected.shape[1]])
	plt.xticks(range(X_train_selected.shape[1]), [feature_names[i] for i in indices if selected_features[i]], rotation=45)
	plt.tight_layout()
	plt.show()
\end{lstlisting}


Upon execution of the revised script, we observed a substantial improvement in the model's performance. The accuracy score increased from the earlier \textbf{ 59.27\% to 73.32\%}, representing a significant enhancement of 14 percentage points. This marked improvement demonstrates the efficacy of the implemented strategies in optimizing the classifier's performance. 
These results not only validate our approach but also highlight the potential for further optimization in similar classification problem. Certainly other strategies could be employed to continue optimizing the performance of this classifier. After the classifier meets our expected result, the classifier could be saved and be deployed to fulfill the automatic classification task of word duration for a new dataset of spontaneous speech given the features such as  word length, word frequency, PhraseLength are taken. 

\subsubsection{Support vector machine}
SVM is a powerful and versatile supervised machine learning model, particularly well-suited for classification tasks \citep{hearst1998support}. It works by finding the hyperplane that best divides a dataset into classes with the maximum margin, i.e., the maximum distance between data points of both classes. SVMs are effective in high-dimensional spaces and relatively immune to overfitting, especially in cases where the number of dimensions exceeds the number of samples.


The following Python script illustrates the implementation of a SVM for classifying word speech durations into different categories. This code exemplifies the step-by-step process of developing and deploying a classification model, specifically tailored to predict duration ranges of words in the \texttt{Buckeye corpus}. Each stage of the implementation corresponds to the machine learning steps outlined in Table~\ref{table:ml_steps}. 

\begin{lstlisting}*[caption={Python code for loading data and training a SVM classifier}, label=lst:svm]
	import pandas as pd
	import numpy as np
	from sklearn.model_selection import train_test_split
	from sklearn.svm import SVC
	from sklearn.metrics import accuracy_score
	
	# Feature selection
	X = df[['CiteLength', 'PhraseLength', 'PhraseRate', 'Deletions', 'WordLength', `LogWordFrequency']]
	y = df[`range_label']
	X = X.dropna()
	y = y.loc[X.index]
	
	X_train, X_test, y_train, y_test = train_test_split(X, y, test_size=0.25, random_state=42)
	
	# Create and train the Support Vector Machine Classifier
	clf = SVC(kernel=`linear', C=1.0, random_state=42)
	clf.fit(X_train, y_train)
	
	preds_svm = clf.predict(X_test)
	accuracy = accuracy_score(y_test, preds_svm)
	print(``Accuracy:", accuracy)
	#accuracy 51.2334%
\end{lstlisting}

We used the SVC from the package ``scikit-learn'' to create a classifier. kernel=`linear' specifies that we are using a linear kernel for the SVM. ``C=1.0'' is the regularization parameter, which controls the trade-off between maximizing the margin and minimizing the classification error. We fit the classifier to the training data using the fit() method. Then, we make predictions on the test data using the predict() method. Finally, we calculate the accuracy of the model using ``accuracy\_score()'' from scikit-learn. We can adjust the kernel type (linear, rbf, poly, etc.) and other hyperparameters of the SVM as needed based on the specific requirements and the characteristics of the dataset. The final accuracy reaches about 51\%, which is close to the one predicted by the random forest. The consistent accuracy results show that these factors play stable roles in machine learning. 
We selected four features to visualize how SVM helps classify them.


Following the RF analysis, we streamlined the factors in our training dataset. For example, using only `WordLength' and `WordFrequency,' we achieved a prediction accuracy of 50.92\%. In contrast, CiteLength' and PhraseRate' resulted in a lower accuracy of 38.36\%, while combining `Deletions' with `WordLength' reached 42.72\%. Such results are quite close to those run in the random forest. This approach helps us understand that `WordLength' and `WordFrequency' are more critical for improving performance. In the following statistical analysis, we should treat both factors as control predictors. The addition of the features, such as `Age' and `Speaker', improved the accuracy scores, and the scores are almost similar to the ones in the RF experiments (i.e., 53.37\%, 59.42\%).  


Despite the initial implementation, optimizing the SVM classifier is a critical step in improving the performance of SVM classifier. Several strategies can be employed to enhance the classifier's accuracy and efficiency. 
Following the optimization of the SVM classifier, its accuracy score improved to 72.33\%, which is comparable to the performance of the optimized Random Forest classifier. The specific strategies and their corresponding implementation script are detailed in \textbf{Appendix B}. These findings highlight that optimizing machine learning implementation is crucial for enhancing model performance. An optimized ML model can then be deployed to process new datasets to make predictions effectively. 

\subsection{Study 2: Statistical analysis}

Statistical analysis is broadly divided into two main types: descriptive statistics and inferential statistics. Descriptive statistics summarizes and describes the features of a dataset through metrics like \texttt{mean}, \texttt{median}, and \texttt{standard deviation}, etc., helping to visualize and understand data distributions. Inferential statistics, on the other hand, use samples of data to make generalizations or predictions about a larger population, employing techniques such as \texttt{hypothesis testing}, \texttt{regression analysis}, and \texttt{confidence intervals}. This distinction allows statisticians and researchers to both understand the data they have and to infer properties about data they do not have, supporting decision-making across fields like economics, medicine, engineering, and social sciences. 

We first employed descriptive correlation to explore in the data of the Buckeye corpus, and then applied some advanced regression models to make further explorations. Implementing advanced regression models for statistical analysis is significantly distinct from the implementation of ML models. 

\subsubsection{Correlation}
\begin{figure}
	\vskip 0.2in
	\begin{center}
		\centerline{\includegraphics[width=0.76\textwidth]{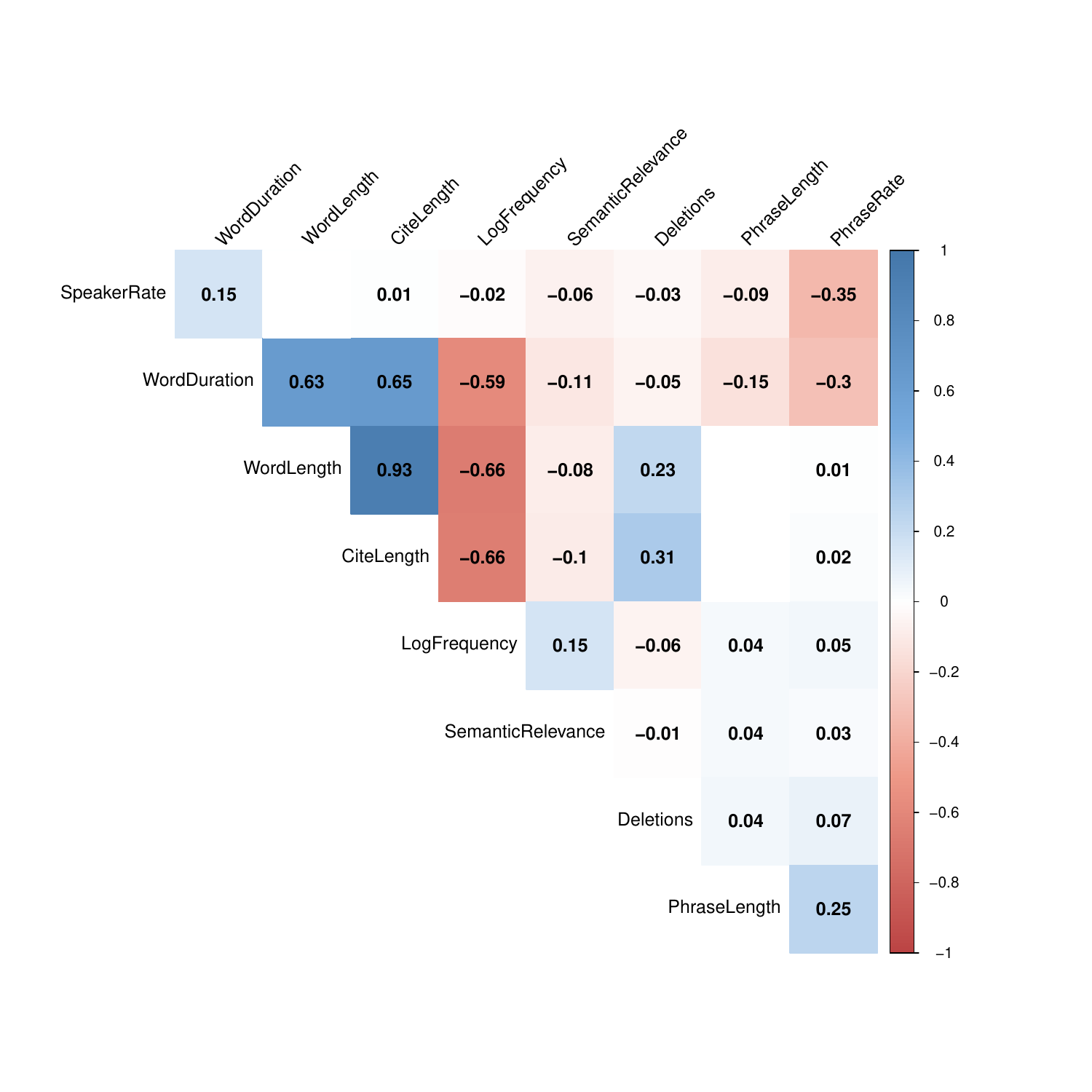}}
		\caption{Correlations among various factors}
		\label{corr_fig2}
	\end{center}
	\vskip -0.26in
\end{figure}

The first statistical analysis is to explore the Pearson's correlationship among these factors we have applied. Using in-built \texttt{R} function, we could obtain the correlation matrix for these variables, and it highlights significant relationships among various features, as shown in Fig. \ref{corr_fig2}. Specifically, a very strong positive correlation exists between ``WordLength'' and ``CiteLength'' ($\rho$ = 0.926), indicating that as words get longer, they generally contain more syllables. This is complemented by strong negative correlations, such as between ``WordLength'' and ``LogFrequency'' ($\rho$ = -0.662), where longer words appear less frequently. Similarly, ``WordDuration'' and ``LogFrequency'' show a moderate negative correlation ($\rho$ = -0.589), suggesting that less frequent words tend to have longer durations. These patterns suggest a close link between the physical characteristics of words and their usage frequencies.

Further analysis shows relationships affecting phrase dynamics and overall speech patterns. Here we included more factors, such as PhraseLength (the alphabet number of phrase where the target word is located), SpeakerRate (the average speaking speed for this speaker) to show more divergent correlations. For instance, there is a moderate positive correlation between ``PhraseRate'' and ``PhraseLength'' ($\rho$ = 0.249), indicating that longer phrases typically have a faster speech rate. Conversely, a moderate negative correlation between ``SpeakerRate'' and ``PhraseRate'' ($\rho$ = -0.349) suggests that faster phrase rates correlate with slower overall speaking rates, potentially reflecting variations in speaking dynamics based on phrase complexity. These findings highlight the specific ways in which speech elements interact, providing valuable insights for studies in linguistic patterns and speech processing. Importantly, while these correlations reveal trends, they do not imply causation, and correlations above approximately 0.5 are particularly noteworthy in social science contexts.

While correlation analysis provides a useful descriptive measure of the relationship between two given factors, it has limitations when exploring complex interactions and relationships. Advanced regression models are necessary to reveal more unique and multifaceted relationships among variables. For instance, as we know, a response variable is supposed to be influenced by a number of factors. When we aim to understand how multiple factors collectively influence this response variable, or when we need to account for random effects, correlation analysis alone is insufficient. Advanced regression models offer other insights: non-linear relationships, random effects, controlling for confounders etc. For example, in language production research, we might use a mixed-effect regression model to examine how factors such as age, gender, word frequency, and speaking rate collectively influence speech duration, while also accounting for random effects of individual speakers and words. This approach would provide a much richer understanding of the complex relationships at play compared to simple correlation analysis. 

\subsubsection{Mixed-effect regression analysis}

Regression models are fundamental tools in statistical analysis, serving multiple crucial purposes in understanding and quantifying relationships between variables. At their core, these models aim to elucidate how changes in one or more independent variables affect a dependent variable. 
Regression models enable precise estimation of how much the dependent variable is expected to change when an independent variable is altered, holding other factors constant. Furthermore, regression models are instrumental in controlling for confounding variables, thereby isolating the effects of specific predictors and providing a clearer picture of complex relationships. This function allows researchers and analysts to test hypotheses, and identify the most influential factors in complex systems.

Regression models in statistical analysis encompass a wide range of techniques, from simple linear regression to more complex generalized mixed-effect models. These include ordinary least squares (OLS), logistic regression, polynomial regression, and advanced mixed-effect models like Linear Mixed-Effects Regression (LMER) and Generalized Additive Mixed Models (GAMM). LMER and GAMM are particularly powerful for analyzing hierarchical or nested data structures and handling non-linear relationships, respectively. Simple linear regression and mixed-effect regression models differ significantly in their structure and capabilities. Simple linear regression models the relationship between one dependent variable and one independent variable, assuming a linear relationship and independence of observations. In contrast, mixed-effect regression models can incorporate multiple independent variables and account for both fixed effects (population-level trends) and random effects (group-level variations). This makes them particularly useful for analyzing hierarchical or nested data structures, such as repeated measures or longitudinal studies. Mixed-effect models allow for different intercepts and/or slopes for different groups within the data, providing a more nuanced understanding of relationships that may vary across subgroups. 
These mixed-effect models offer significant advantages over ordinary regression models by accounting for both fixed and random effects, allowing for the analysis of grouped or longitudinal data, and providing more accurate estimates when dealing with correlated observations. 


Table~\ref{table:gamm_steps} summarizes the steps for implementing and analyzing the relationship between a response variable and independent variables using mixed-effect regression models.

\begin{table}[!ht]
	\centering
	\caption{Steps for implementing and analyzing statistical regression models}
	\begin{tabular}{|c|l|p{7cm}|}
		\hline
		\textbf{Step} & \textbf{Name} & \textbf{Description} \\
		\hline
		1 & Data preparation & Ensure data is clean, properly formatted; check for missing values and outliers; explore data distribution and relationships. \\
		\hline
		2 & Model specification & Define response variable; identify the factor of the interest;include the control predictors; determine random effects structure; choose appropriate smoothing functions. \\
		\hline
		3 & Model fitting & Use functions like \texttt{gamm()} or \texttt{bam()}; specify model formula; choose distribution family and link function. \\
		\hline
		4 & Model evaluation & Check convergence; examine summary statistics; assess model fit; perform model diagnostics. \\
		\hline
		5 & Model comparison & Fit alternative models; compare using \texttt{AIC} to understand the contribution of variable (other evaluation criteria: P-values, Confidence Intervals etc.). \\
		\hline
		6 & Interpretation & Examine smooth term plots; interpret parametric coefficients; analyze random effects structure. \\
		\hline
		7 & Validation & Use cross-validation techniques; check for overfitting. \\
		\hline
		8 & Visualization & Create plots of smooth terms and interactions; visualize model predictions and confidence intervals. \\
		\hline
		9 & Reporting & Summarize key findings and model performance; present relevant plots and tables. \\
		\hline
		10 & Iteration & Refine model based on results and domain knowledge; consider additional predictors or alternative structures if necessary. \\
		\hline
	\end{tabular}
	\label{table:gamm_steps}
\end{table}

\vspace{0.2cm}
\indent \textbf{1) LMERs}: 

The first model is Linear Mixed-Effects Regression (\textbf{LMER}) (\citealp{bates2010lme4}; \citealp{kuznetsova2017lmertest}). LMER is a statistical model used to analyze data with both fixed effects, which represent the main variables of interest expected to have a consistent impact across the dataset, and random effects, which account for inherent variations from grouped or nested data sources. This model is particularly useful for dealing with hierarchical structures, such as students within classrooms, or data from repeated measures on the same subjects, allowing for more accurate estimates by accounting for both within-group and between-group variability. LMER helps in understanding complex datasets by modeling the dependencies and structure within the data, making it a powerful tool for robust statistical analysis. 

The present study employed LMER to explore how independent variables and random variable take effects on the dependent variable. Our primary question is to explore whether some variables (i.e., semantic relevance) take effect on word duration. This question is clearly distinct from the machine learning models to make classification in Study 1. After knowing the research questions and the factors in the data, we select LMER to build up the setups, which could be followed by Table~\ref{table:gamm_steps}. The dependent variable is ``word duration'', and independent variables include word length, word frequency, semantic relevance etc., and the random variables (i.e. Speaker, Age and Sex). The data number (n) in the LMER fittings is 262342.  

A typical mixed-effect regression model is consisting of two parts: dependent variable and independent variables. The independent variables are typically composed of three (or four) elements: control predictors, factors of our interest, and random variables, and probably with other predictors. The basic structure of LMER fittings is formulated as follows. Control predictors such as word length and word frequency were included. For instance, longer words tend to have longer durations, whereas more common words typically result in shorter durations \citep{wright1979duration}. Random variables such as age, sex and speaker could be included. The variable of interest in our study is ``semantic relevance,'' which is determined by the research question and prior investigations.  A significant body of research has demonstrated that other various factors influence word duration in speech. For example, words with more syllables generally have longer durations, a phenomenon closely related to word length. Faster speech rates are associated with shorter word durations, a relationship confirmed in multiple studies (\citealp{fujisaki1997prosody}; \citealp{darling2021speech}; \citealp{haley2023normative}). Furthermore, words that experience greater phonological simplification often exhibit shorter durations, a trend noted in research on conversational language (\citealp{yaruss2000converting}; \citealp{baker2009variability}; \citealp{waage2024effect}). These factors, along with control predictors and the predictor of interest, should jointly influence word duration. Therefore, they should be incorporated into statistical regression models. However, it remains unclear whether the semantic relationship between a target word and its context impacts word duration. This knowledge gap is the key motivation for our investigation, and we plan to use mixed-effect regression models to explore this relationship.

\begin{equation*}
	\begin{aligned}
		\textcolor{purple}{\text{Dependent Variable}} \sim {} & \framebox{$\textcolor{blue}{(\text{Control Predictor}_1 + \text{Control Predictor}_2 + \cdots)}$} \\[0.5em]
		& + \framebox{$\textcolor{red}{(\text{Factor of Interest})}$} \\[0.5em]
		& + \framebox{$\textcolor{red}{(\text{Other Predictors (Optional)})}$} \\[0.5em]
		& + \framebox{$\textcolor{green}{(\text{Random Variable}_1 + \text{Random Variable}_2)}$}
	\end{aligned}
\end{equation*}



Once the setup of the mixed-effect fitting is complete, we can consider various \texttt{R} packages for implementation, such as ``lme4'', ``nlme'', and ``glmmTMB''. These packages do not require training themselves but provide the necessary tools for fitting and analyzing mixed-effect models. Understanding this is crucial for appreciating the significance of performance indicators, with the Akaike Information Criterion (AIC) emerging as the preferred evaluation criterion for model comparison because AIC is an estimator of prediction error and thereby relative quality of statistical models for a given set of data. The AIC balances model fit against complexity, with smaller values indicating better performance, as illustrated in Table \ref{lmeraic}. When interpreting AIC, we consider that differences of less than 2 may suggest similar levels of support for the models being compared. Additionally, while AIC is valuable, it should not be used in isolation; other factors such as theoretical grounding, research questions, and result interpretability should also inform model selection. 

{\scriptsize{
		\begin{itemize}
			
			\item lm1=lmer(WordDuration$\sim$WordLength+LogWordFreq+CiteLength+PhraseRate+(1|Sex)+(1|Speaker))
			
			\item lm2=lmer(WordDuration$\sim$WordLength+LogWordFreq+CiteLength+SemanticRelevance+PhraseRate+\\Deletions+(1|Age)+(1|Sex)+(1|Speaker))
			
			\item lm3=lmer(WordDuration$\sim$WordLength+LogWordFreq+CiteLength+SemanticRelevance+PhraseRate+\\Deletions+(1|Age)+(1|Speaker))
			
			\item lm4=lmer(WordDuration$\sim$WordLength+LogWordFreq+CiteLength+SemanticRelevance+PhraseRate+\\(1|Age)+(1|Speaker))
			
			\item lm5=lmer(WordDuration$\sim$WordLength+LogWordFreq+CiteLength+SemanticRelevance+PhraseRate+\\Deletions+(1|Sex)+(1|Speaker))
			
\end{itemize}}}

\begin{table}[ht]
	\centering
	\caption{LMER Model Comparison with AIC Values (n=262342)}
	\begin{tabular}{@{}ccc@{}}
		\toprule
		Model & Degrees of Freedom (df) & AIC        \\ \midrule
		lm1   & 8                        & -433759.4  \\
		lm2   & 11                       & -459522.0  \\
		lm3   & 10                       & -459523.8  \\
		lm4   & 9                        & -433952.1  \\
		lm5   & 10                       & -459523.8  \\ \bottomrule
	\end{tabular}
	\label{lmeraic}
\end{table}

After running the LMER fittings, we evaluated and interpret the results. First, most of factors listed were significant in these fittings (the threshold of \textit{p}-value smaller than 0.01). In other words, these factors significantly influenced word duration. The detail of interpreting the fitting result is seen in\textbf{Appendix C}. Despite this, clearly, the smaller AIC indicates better performance.  ``lm3'' / ``lm5”''has the smallest AIC, and this suggests that these factors and the random variable have significant effects on word duration.  Compared with ``lm3'' and ``lm4'', we find that the factor ``Deletion'' did  significantly contribute to AIC in the model. However, when the random effect ``age'' or ``sex'' or both appears, they have no great difference. Through comparison, we found that the random factor ``sex'' had no significant effect on word duration, which is consistent with the feature selection (`speaker') in traning ML models. These indicate that features' selection play a crucial role in both machine learning and stastistical regression analysis. 

Additionally, these control predictors are significant across all cases. Previous research has demonstrated the importance of control variables such as ``WordLength'', ``LogWordFreq'', ``CiteLength'', and ``PhraseRate''. Our focus is on whether the new variable, semantic relevance, affects word duration. The results from the LMERs consistently show that ``semantic relevance'' influences word duration, as the variable is significant in all fittings. These findings indicate that the degree of semantic connection between the contextual information and the target word (measured by ``semantic relevance'') significantly impacts word duration in spontaneous speech, a form of language production. This suggests that semantic relevance plays a role in language production processes.

\vspace{0.2cm}
\indent \textbf{2) GAMMs:}

Next, we applied Generalized Additive Mixed Models (\textbf{GAMM}) in analyzing these factors in a similar way. GAMMs are an extension of Generalized Linear Mixed Models, incorporating non-linear relationships between the dependent and independent variables through smooth functions \citep{wood2017generalized}. GAMMs allow for both fixed and random effects, accommodating complex variations within hierarchical data structures. The ``additive'' part of GAMM means that the model expresses the dependent variable as a sum of smooth functions of predictors, along with any random effects and an error term. This flexibility makes GAMMs particularly useful for modeling non-linear trends in data, where the effect of variables is not strictly linear and may vary by group or over time. Using GAMM with the same setups could help cross-verify the results obtained from the LMER fittings. The general mathematical equation for a GAMM is:

\begin{equation}
	g(E(Y_i)) = X_i\boldsymbol{\beta} + f_1(x_{1i}) + f_2(x_{2i}) + \cdots + f_m(x_{mi}) + Z_i\mathbf{b}_i + \epsilon_i
\end{equation}

Where $g(\cdot)$ is the link function; $E(Y_i)$ is the expected value of the response variable for the $i$-th observation; $X_i\boldsymbol{\beta}$ represents the parametric fixed effects, where $X_i$ is the $i$-th row of the model matrix for parametric terms and $\boldsymbol{\beta}$ is the vector of fixed effect coefficients; $f_j(x_{ji})$ are smooth functions of covariates $x_j$; $Z_i\mathbf{b}_i$ represents the random effects, where $Z_i$ is the $i$-th row of the random effects model matrix and $\mathbf{b}_i$ is the vector of random effects; $\epsilon_i$ is the error term. 

The smooth functions $f_j(\cdot)$ are typically represented using basis expansions, such as splines. The random effects $\mathbf{b}_i$ are usually assumed to follow a normal distribution with mean zero and some covariance structure. This equation combines elements of generalized linear models, additive models, and mixed-effects models, allowing for flexible modeling of non-linear relationships while accounting for hierarchical data. Compared with Equation (3), this GAMM equation (4) is basically consistent with the math equation of regression model in machine learning. More details on their differences could be seen in the Discussion Section 4.2.

Implementing a GAMM follows a similar procedure to fitting a LMER fitting, requiring the setup of both dependent and independent variables. The independent variables include control predictors, the factors of our interest and the random variables, which are identical with the ones in LMERs. After running the GAMM fittings, we need to evaluate and interpret the results. We listed a number of GAMM fittings and made comparison by referring to AIC. The biggest differences between GAMM and LMER is that GAMM could employ the function \texttt{s()} and interaction smooth \texttt{te()}. The smooth function better gets model fittings for some factors, and the interaction smooth could find the interaction among some given factors. The independent and dependent variables were set similarly as the ones in LMER. The AIC results are shown in Table \ref{gamaic}.  

{\scriptsize{
		\begin{itemize}
			
			\item t1=bam(WordDuration$\sim$s(WordLength)+s(LogWordFreq)+s(CiteLength)+s(SemanticRelevance)+\\s(PhraseRate)+s(Deletions)+s(Age, bs="re")+s(Speaker, bs="re"))
			
			\item t2=bam(WordDuration$\sim$s(WordLength)+s(LogWordFreq)+s(CiteLength)+s(SemanticRelevance)+\\s(PhraseRate)+s(Deletions)+s(Sex, bs="re")+s(Speaker, bs="re"))
			
			\item t3=bam(WordDuration$\sim$te(WordLength,LogWordFreq)+s(CiteLength)+s(SemanticRelevance)+\\s(PhraseRate)+s(Deletions)+s(Speaker, bs="re"))
			
			\item t4=bam(WordDuration$\sim$te(WordLength,LogWordFreq)+s(CiteLength)+s(PhraseRate)+s(Deletions)+\\s(Sex, bs="re")+s(Speaker, bs="re"))
			
			\item t5=bam(WordDuration$\sim$te(WordLength,LogWordFreq)+s(PhraseRate)+s(Deletions)+s(Sex, bs="re")+\\s(Speaker, bs="re"))
			
			\item t6=bam(WordDuration$\sim$s(WordLenght)+s(LogWordFreq)+s(CiteLength)+s(PhraseRate)+s(Deletions)\\+s(Sex, bs="re")+s(Speaker, bs="re"))
			
\end{itemize}}}


\begin{table}[ht]
	\centering
	\caption{GAMM Model Comparison with AIC Values (n=262342)}
	\begin{tabular}{@{}ccc@{}}
		\toprule
		Model & Degrees of Freedom (df) & AIC        \\ \midrule
		t1    & 281.8724                & -486185.2  \\
		t2    & 281.7350                & -486185.1  \\
		t3    & 282.5285                & -486017.5  \\
		t4    & 279.2962                & -485994.0  \\
		t5    & 275.9720                & -471297.8  \\
		t6    & 278.3133                & -486159.7  \\ \bottomrule
	\end{tabular}
	\label{gamaic}
\end{table}

Similarly, most of these factors are significant in these cases, which is consistent with the results in LMER fittings. Random effect plays a crucial role in regression analysis \citep{baayen2017cave}. Interpreting the parameters in GAMM fitting result is seen in \textbf{Appendix C}. Through the analysis, we found that the random variable ``age'' is not significant in ``t1'' (\textit{p}-value is smaller than 0.01), which is consistent with the analysis of LMER fittings and feature selections in training ML models. T2 is the best fitting compared with other ones. Comparing ``t2'' and ''t6'', we could find that the factor ``semantic relevance'' substantially contributed to the model, namely, the factor  of our interest, ``semantic relevance'',  had a significant impact on word duration. The result is consistent with the one from the LMER fittings.

\begin{figure}
	\vskip 0.2in
	\begin{center}
		\centerline{\includegraphics[width=1.18\columnwidth]{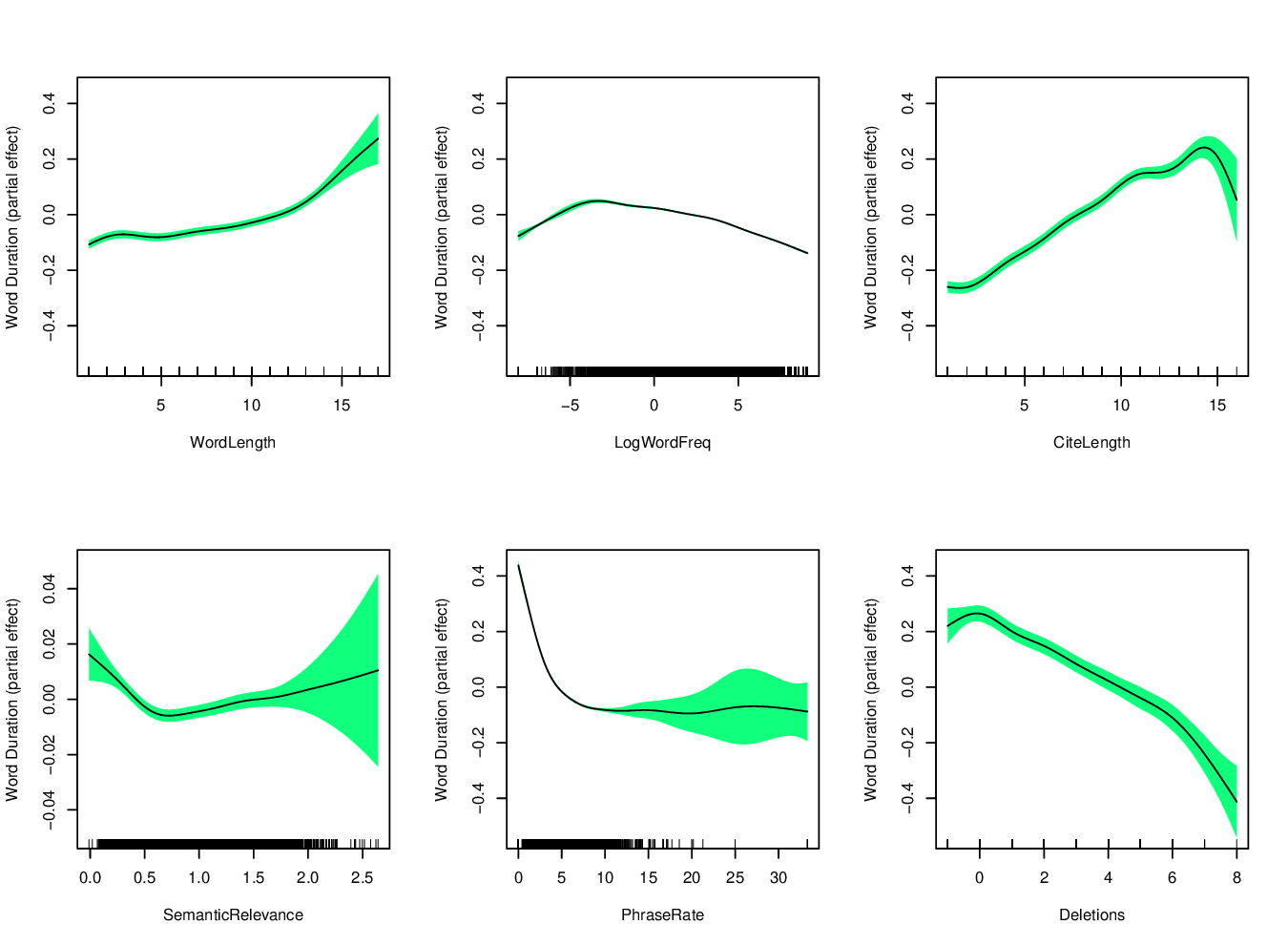}}
		\caption{Partial effects of a given factor on word duration. (Note: The \textit{x-axis} signifies the metrics, while the \textit{y-axis} delineates word duration. The plots are arranged in two rows, each containing three graphs. The top row, from left to right, displays the predictors: word length, word frequency, and CiteLength. The bottom row, also from left to right, shows: semantic relevance, phrase rate, and deletions. Each curve for an individual plot visually articulates the relation between a predictor variable and the response variable. A steeper incline on these curves demonstrates a more robust impact between the predictor and word duration, whereas gentler slopes imply a less pronounced effect. Moreover, when a curve fluctuates around zero, its effect vanishes. The  \textit{p}-value is smaller than 0.0001 in each plot.)}
		\label{gam_fig3}
	\end{center}
	\vskip -0.26in
\end{figure}

To further understand the impacts of these factors, we visualized the partial effects of various predictors on word duration, as shown in Fig.~\ref{gam_fig3}. The visualization of partial effects in GAMM results is a valuable tool to make a better understanding of the effects of the included predictors. In addition to \textit{p}-values and AIC values, the curve trend of partial effect for a given predictor could reveal significant information. The instruction on how to read the curve of partial effects could be seen in the caption of Fig.~\ref{gam_fig3}. According to Fig.~\ref{gam_fig3}, we found that word length and ``CiteLength'' are negatively correlated with word duration, indicating that as these factors increase, the word duration decreases. It is understandable that words with more syllables and letters could be spoken more slowly. Conversely, fast phrase rates and more deletions lead to shorter word durations. Word frequency and semantic relevance exhibit complex effects. When word frequency is low (less than 2), it is positively correlated with word duration; however, when word frequency exceeds 3, its impact turns negative. Similarly, when a word is less semantically related to the context (score less than 2), it negatively affects word duration. In contrast, high semantic relevance (score greater than 2) positively influences word durations.

Semantic relevance measure quantifies the degree to which contextual information semantically influences the target word, providing a unique understanding of language processing. The results of our GAMM fittings reveal that semantic relevance significantly impacts word duration in speech, suggesting that words with higher semantic relevance to their context are produced differently than those with lower relevance. This finding opens up exciting avenues for further research into the mechanisms of language production. For instance, researchers could explore how varying levels of semantic relevance affect not only word duration but also other aspects of speech, such as intonation, stress patterns, or articulatory precision. Additionally, investigating how semantic relevance interacts with other linguistic factors, such as word frequency, could provide a more comprehensive model of language production. Recent research has demonstrated the efficacy of semantic relevance in predicting eye movements during reading across multiple languages (\citealp{sun2023reading}; \citealp{sun2023optimizing}; \citealp{sun2024eeg}). Given that reading is a quintessential form of language comprehension, this finding, combined with its impact on speech production, establishes semantic relevance as a robust predictor in both language comprehension and production processes. Consequently, this metric appears to play a significant role in human language processing, bridging the gap between how humans comprehend and produce language. The broad applicability of ``semantic relevance'' suggests its fundamental importance in language processing and opens up new avenues for exploring the intricate cognitive mechanisms.

In summary, the LMERs and GAMMs can cross-verify each other, consistently confirming the effects of these predictors and random variables across the two different models. Our main concern on the influence of ``semantic relevance'' on word duration could be addressed well: the metric is significant. Further analysis could explore how semantic context affects the processing of target words during language production, which can be examined from both psychological and linguistic perspectives. This underscores the importance of statistical analysis and its interpretations in the fields of language science and cognitive studies. 

\section{Discussion}

In summary, we employed machine learning methods such as RFs and SVMs to train on the training dataset, allowing them to identify patterns that were then applied to predict the classification of various word durations in the test dataset. In contrast, statistical methods like LMER or GAMM were used primarily to analyze how specific factors affect word duration. These statistical methods do not require training on a testing dataset to determine necessary model parameters, and they are not typically involved in pattern recognition. 
This section provides a deeper analysis of the differences observed in our experiments, with a focus on the regression models used in both machine learning and statistical analysis. Additionally, we discuss how machine learning and statistical approaches can be applied differently in the context of cognitive science.

\subsection{Key differences between machine learning and statistics}

Based on Table~\ref{table:ml_steps}, Table~\ref{table:gamm_steps}, and (a) of Table~\ref{tab:ml_stats_differences} as well as the Python scripts and R scripts, we overview the differences on implementation of machine learning and regression statistical analysis. The implementation of machine learning models and statistical regression analyses, while sharing some common ground, exhibit distinct approaches, implementations, and emphases. Machine learning typically follows a more iterative, performance-driven process, focusing on predictive accuracy and model optimization. 

Machine learning begins with problem definition and data gathering, followed by preprocessing and feature selection. The process then moves through model selection, data splitting, training, and evaluation, with a strong emphasis on tuning and optimizing model performance using metrics like precision, recall, and F1-score. In contrast, statistical regression analysis, particularly when implementing advanced models like GAMMs, adopts a more structured, inference-oriented approach. It starts with meticulous data preparation and exploratory analysis, followed by careful model specification that considers factors of interest, control predictors, and random variables. The analysis then progresses through model fitting, evaluation, and comparison, with a focus on statistical significance like \textit{p}-value, confidence intervals, and information criteria like AIC. Statistical approaches place greater emphasis on model interpretation, validation of assumptions, and the iterative refinement of models based on both statistical results and domain knowledge. While both methodologies involve data preparation, model fitting, and evaluation, statistical regression analysis typically requires more upfront theoretical consideration and places higher importance on explaining relationships and testing hypotheses, rather than solely optimizing predictive performance.

Let us correlate our experiments with the key further differences outlined in (b) of Table~\ref{tab:ml_stats_differences} to illustrate their specific applications in research on language and cognitive sciences. First, random forest and SVM were primarily employed to predict varying ranges of word durations in a new dataset. It was observed that both methods achieved a prediction accuracy of 73\% for classifying word durations. However, correlation analyses and mixed-effects regression demonstrated are not required to predict word durations on new data. The reason for this is that ML and statistics are taken to solve different questions: ML is to make predictions for clustering, classification or regression; in contrast, statistics aims to explore the relationship among variables instead of making predictions for new data. Second, machine learning methods often prioritize model accuracy over the interpretability of the extracted features. While the features used in these methods are interpretable, real-world ML algorithms focus on feature effectiveness in prediction. In contrast, features (i.e., variables or factors) in statistical analysis should be interpretable from linguistic or cognitive perspectives. In fact, all the factors we selected are highly interpretable. Third, ML focuses on how features contribute to the prediction accuracy of word duration (e.g. the feature ``Speaker''). Statistical regression models, however, aim to determine how certain factors affect word duration and the strength of these effects, often within the framework of established statistical theories and constraints. For instance, GAMMs or LMERs provide insights into how various factors influence model performance, thereby helping researchers understand the significant impact of these factors on the dependent variable (i.e., word durations). 

\subsection{Differences between regression models used in ML and statistics}

Section \ref{exem} and the LMER/GAMM implementations employed regression models for statistical analysis. It is important to address the distinctions between using regression models in ML and in statistical analysis, as these differences often cause confusion. While regression models in both fields may use the same algorithms, they differ significantly in terms of purpose, methodology, and application. 
These differences stem from the distinct objectives and methodologies of machine learning and traditional statistical analysis.

ML regression models, such as linear regression, decision trees, or more complex ensemble methods like decison trees in Random Forest, primarily focus on prediction. They excel at capturing complex relationships in large datasets and are often used when the primary goal is to make accurate predictions on new, unseen data. The section\ref{exem} presented an example of using a regression model to predict children's cognitive development based on early linguistic behaviors. Regression models in ML are well-suited for making future predictions. In this case, regression models are an appropriate choice for such a predictive task. Going back to Equation (3), the training and optimization of the regression model in the context of machine learning aim to find the best set of optimized parameters for this equation. These models typically employ strategies like regularization to prevent overfitting and use cross-validation for model selection and performance evaluation. However, ML regression models often sacrifice interpretability for predictive power. In other words, interpretability is not required for these models when they are applied to predictions.

In contrast, for instance, GAMMs, which fall under the umbrella of statistical analysis, strike a balance between flexibility and interpretability. They extend generalized linear mixed models by incorporating smooth functions to model non-linear relationships between predictors (independent variables) and the response variable (dependent variable). This way, GAMMs are particularly useful when the goal is understanding and interpreting the relationships between variables. The section\ref{exem} mentioned that regression statistical analysis is taken to explore relationships among the children cognitive development and linguistic variables (e.g., syntactic complexity, vocabulary size/diversity, sentence length, speech rate or fluency measures etc.), providing comprehensive insights into the language development trends or characteristics. 

Now referring back to Equation (4), we understand that a GAMM needs a process of parameter estimation in this equation, that is, ``fitting'', which involves estimating fixed effect coefficients, random effect variances, and smoothing parameters for non-linear functions. Techniques like Maximum Likelihood Estimation (MLE), Restricted Maximum Likelihood (REML), or Penalized Likelihood Estimation are typically employed, often using iterative algorithms such as Penalized Iteratively Reweighted Least Squares (P-IRLS). The goal is to accurately describe the relationships between variables, with a focus on model interpretability. In contrast, a regression model in ML also needs such a process but it is termed ``training'' (see Equation (3)). The training process in ML often involves minimizing a loss function--commonly via optimization algorithms such as Gradient Descent, Stochastic Gradient Descent (SGD), or more advanced variants like Adam or RMSprop.   

Despite their similarities in parameter estimation, GAMMs and machine learning regression models serve fundamentally different purposes. The goal of fitting a GAMM is not to achieve the best predictive performance but to gain insights into the underlying patterns and structures in the data because understanding variable interactions is more important than prediction. Regression models in ML, on the other hand, are primarily employed for making accurate predictions. The training process in ML is focused on optimizing predictive performance, and the goal is to create a model that generalizes well to unseen data. Moreover, the terminology and methodology between GAMMs and machine learning regression models reflect their differing objectives. GAMM fitting focuses on model interpretation, balancing complexity with interpretability through the use of smoothing parameters. Overfitting is controlled by adjusting these parameters, with the aim of understanding the significance of the relationships between variables. Machine learning training, in contrast, emphasizes predictive accuracy. Models are trained to minimize error and maximize generalization, often using advanced optimization and tuning techniques. The complexity of the model is balanced by validation techniques to ensure it performs well on new, unseen data. Additionally, a trained ML model is typically tested on a separate test dataset, whereas a fitted statistical model does not require testing on a split dataset. This difference arises from their distinct objectives: ML models aim to predict outcomes, while statistical models focus on exploring relationships among variables.

GAMMs are often handled by specialized statistical package/software, which may obscure the complexity of the estimation process, while ML models popularize the term ``training'', making the process more explicit in modern data science workflows. However, it is important to recognize that both approaches involve sophisticated methods for learning patterns from data, even if their goals and some techniques differ


One key advantage of GAMMs over machine learning regression models is their ability to provide clear visualizations of smooth effects and offer statistical inference on model components. This makes GAMMs particularly valuable in research settings where understanding the nature of relationships and testing specific hypotheses are crucial. GAMMs also handle uncertainty more explicitly, providing standard errors (e.g., AIC) and confidence intervals (e.g., \textit{p}-value,) for model components.
On the other hand, the training process in ML is focused on optimizing predictive performance, and the goal is to create a model that generalizes well to unseen data. Metrics like mean squared error or R-squared on a held-out test set measure this performance. Overall, while their objectives and methods differ, both approaches use iterative procedures to learn patterns from data, reflecting distinct traditions in data analysis. 

Additionally, the concept of \texttt{shrinkage} in regression models is the tendency for model performance to decrease when applied to new data. Statistical analysis and ML process shrinkage differently. In statistical approaches, shrinkage can be estimated analytically from the model fit itself, based on the model's assumptions. This contrasts with machine learning approaches, which typically determine shrinkage empirically using separate test data or cross-validation. The key point is that statistical models can often predict their own generalization performance without requiring additional data, as long as their underlying assumptions hold true. This approach allows for efficient use of data but relies heavily on the validity of the model's theoretical properties.

In practice, the choice between ML regression models and GAMMs often depends on the specific research questions, the nature of the data, and interpretability. While machine learning models might be preferred for pure prediction tasks or when dealing with extremely complex datasets, GAMMs offer a powerful tool for researchers who need to model non-linear relationships while maintaining interpretability and the ability to perform statistical inference. In some cases, a hybrid approach combining elements of both methodologies might provide the most comprehensive analysis.

\subsection{ML vs. statistics in cognitive science}
So far we have made detailed and consistent comparison between ML and statistical analysis in exploring the Eyebuck corpus. We can conclude that machine learning models are preferable for tasks requiring high accuracy in predictions. Conversely, when the goal is to ascertain relationships between variables or to draw inferences from data, statistical models is more suitable, offering the rigor and transparency needed for such analyses. 

Some might wonder which one is more important in cognitive research. Basically it depends on the research question and tasks. Machine learning and statistics both play crucial roles in cognitive research, and their relative importance depends on the specific research questions and tasks at hand. 

Statistical approaches are typically more important when 1) the primary goal is to understand and explain relationships between variables; 2) testing specific hypotheses about cognitive processes; 3) inferring population parameters from sample data; 4) analyzing experimental data with controlled variables; 5) dealing with smaller datasets where interpretability is key.
In contrast, machine learning approaches become more valuable when 1) the main objective is to predict outcomes or classify pattern; 2) dealing with large, complex datasets (e.g., neuroimaging data); 3) exploring patterns in data without a priori hypotheses; 4) developing models that can generalize to new, unseen data; 5) handling high-dimensional data or non-linear relationships. 

In many cases, a combination of both approaches can provide the most comprehensive insights. For example, statistical methods might be used to formulate and test hypotheses about cognitive processes, while machine learning techniques could be employed to identify complex patterns in neural activity data. Ultimately, the choice between ML and statistical approaches - or the decision to use both - should be guided by the research questions, the nature of the data, and the specific goals of the study. Both methodologies have their strengths, and their complementary use can often lead to more robust and insightful findings in cognitive research. 
The current study is a mixture of machine learning and statistical analysis because it is data-driven research. For instance, in computing the factor \texttt{semantic relevance}, we introduced word embeddings to represent word meanings. Word embedding is a well-known machine learning (specifically, deep learning) technique for deriving word meanings (\citealp{mikolov2013efficient}; \citealp{kusner2015word}; \citealp{ethayarajh2019contextual}), and the algorithm for semantic relevance also incorporates various machine learning methods. However, the computation of semantic relevance remains transparent and interpretable. We employed methods such as GAMM or LMER to investigate how semantic relevance influences word durations. The ultimate goal is to elucidate that semantic relevance could influence human language production; in other words, contextual information plays a vital role when humans produce language. In this sense, GAMM analysis provides interpretable results, further confirming the impact of semantic relevance on the complexity of language production.


While combining approaches can provide more comprehensive insights, practical constraints often lead to focusing on a single dominant method. For instance, in cognitive sciences, publications typically use statistical analysis to explore relationships among variables, followed by further psychological analysis for theoretical interpretations. Although some research mainly on ML algorithms automatically recognizing patterns in datasets, this focus may not align with the main interests in the field of cognitive sciences to some degree. Fortunately, few journals in the related fields such as \textit{Behavior Research Methods} and \textit{Computers in Human Behavior} have published papers emphasizing ML models. Furthermore, with the significantly increasing influence of ML advancements, techniques such as feature extraction, classification, and clustering can greatly be employed in order to enhance the exploration of variable relationships. Nevertheless, employing advanced statistical analysis requires specific skills, including model component setup, evaluation, and comparison. Throughout this process, errors may occur easily, and any resulting misinterpretations can lead to significant theoretical consequences. Overall, as interdisciplinary collaboration grows and computational tools become more accessible, we can expect to see an increase in studies effectively integrating both statistical and machine learning approaches in the future.



\bibliographystyle{apalike}
\bibliography{reference}

\appendix

\section*{ Appendix A: The computation of ``SemanticRelevance'' metric}

\begin{figure}

	\centering
	
	\includegraphics[width=0.86\textwidth]{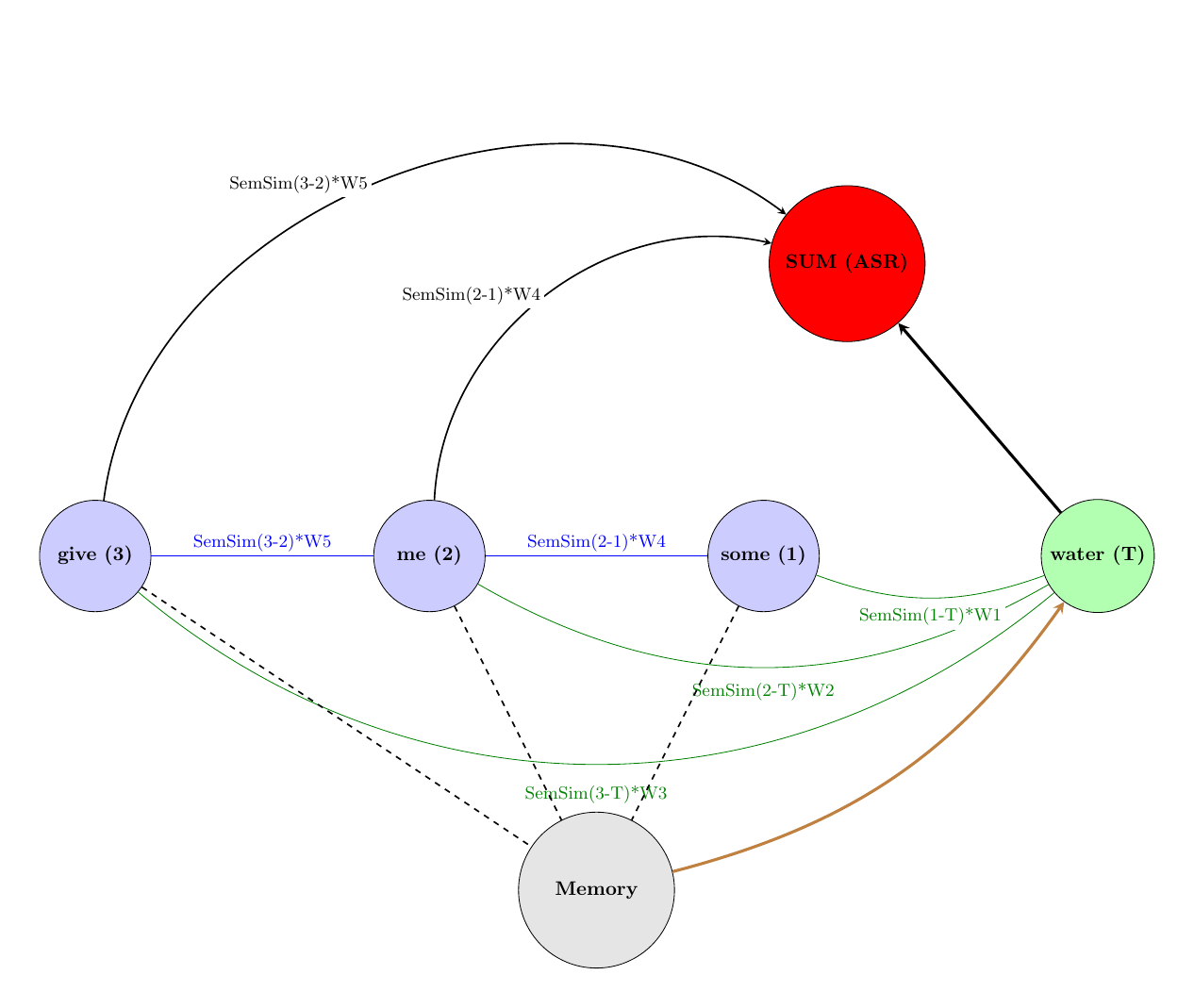}
	
	\caption{The computational method for semantic relevance}
	
	\label{fig:semrev}
	
\end{figure}

The computational method for ``semantic relevance'' can be elucidated by referring to the example depicted in Fig.~\ref{fig:semrev}. The method examines the three preceding words as a context window (for instance, ``give'', ``me'', ``some'') in this example. Inside this window, the semantic relationship between each of these words and the target word (``water'') is assessed. The aggregate of all semantic similarities between the target word and its surrounding words yields a metric of the target word's semantic connection to the broader context. However, this method does not initially consider the differential influence of the surrounding words or the sequence of words. To rectify this, we implement a weighting scheme that is contingent upon the spatial distance between the contextual word and the target word. For example, the significance of the relationship between ``give'' and ``water'' is diminished due to the intervening ``me'', while the relevance between ``some'' and ``water'' is enhanced due to their immediate adjacency. This weighted calculation also takes into account the sequence of words, allowing for a more holistic integration of contextual data. The formula for calculating the attention-aware semantic relevance is expressed as: \( \boxed{\sum SemSim_{(t,\ c)} \cdot W_{(t,\ c)}} \) (where \textit{t} = target word, \textit{c} = surrounding words, \textit{SemSim} = semantic similarity, \textit{W} = weights).  With the example provided, grasping this formula becomes quite clear.

The following details the computational process for calculating various metrics, using the illustration in Figure \ref{fig:semrev}. Each word within the specified window is equipped with a word vector sourced from a pre-trained embedding database. Initially, we determine three semantic similarity scores by comparing the target word to its three preceding words. Additionally, we calculate two semantic similarity scores between any pair of these three preceding words. Subsequently, we apply varying weights to each of the five semantic similarity scores. These weights, which are values between 0 and 1, are assigned based on the proximity of the contextual words to the target word. A closer contextual word receives a higher weight, and conversely, a more distant word gets a lower weight. For instance, the semantic similarity between ``water'' and ``give'' (notated as ``SemSim$_{T-3}$'') is weighted by 1/2, while the similarity between "water" and "some" (notated as ``SemSim$_{T-2}$'') is weighted by 2/3, and the similarity between ``water'' and ``some'' (notated as ``SemSim$_{T-1}$'') is given a full weight of 1. The semantic similarity scores for non-adjacent words, such as ``give'' and ``me'', are reduced; ``SemSim$_{2-3}$'' (the similarity between ``give'' and ``me'') is weighted by 1/3, and ``SemSim$_{1-2}$'' (the similarity between ``me'' and ``some'') is weighted by 1/2. Once each semantic similarity score has been adjusted by its respective weight, we sum these values to derive the "attention-aware semantic relevance" score. This final aggregate represents the overall semantic relevance within the context.

This section clarifies the rationale behind the incorporation of weights in our methodology. The attention-aware methodology used in this research is fundamentally grounded in memory-based processes and has potential relevance to human memory dynamics. As depicted in Figure \ref{fig:semrev}, the preceding words within a window act as a memory stack, reflecting how readers maintain a memory of words they have recently encountered, along with their meanings (with a stack depth of 4 words). The weights applied to measure the metric are contingent upon the proximity of the target word to its neighboring words, with closer words being assigned higher weights and more distant words receiving lower weights. For this study, we have chosen a window size of 4 words, which may mirror the trajectory of a forgetting curve, representing the decrease in memory retention over time, where the retained information is halved each day for several days \citep{loftus1985evaluating}. Within this window stack, words that are closer to the target word are akin to the early days on the forgetting curve, while words that are further away resemble the later days. To emulate the human forgetting mechanism, we have assigned greater weights to words that are closer to the target word and lesser weights to those that are more distant. The application of varying weights is instrumental in synthesizing information regarding the distinct contributions of contextual words, and these weights can also encode information about the sequence of words. 

\section*{ Appendix B: The script of optimized SVM classifier}

Several key strategies are introduced to improve the accuracy of this SVM classifier. First and foremost, feature scaling is implemented to normalize the features and is particularly crucial for SVM performance. This strategy is also applied in optimizing the random forest classifier. This is complemented by a feature selection step using the function of ``SelectKBest'', which identifies and retains only the most relevant features based on ANOVA F-values, potentially reducing noise in the dataset. To address possible class imbalance issues, which are common in many real-world datasets, we incorporate SMOTE (Synthetic Minority Over-sampling Technique). This technique creates synthetic examples of the minority class, helping to balance the dataset and potentially improving the model's ability to learn from underrepresented classes.

Furthermore, we employ an extensive hyperparameter tuning process using GridSearchCV. This systematic approach explores various combinations of important SVM parameters such as the regularization parameter (C), kernel type, gamma, and class weight, to find the optimal configuration for the given dataset. The evaluation process has also been enhanced, now including a comprehensive classification report that provides detailed performance metrics for each class, offering deeper insights into the model's strengths and weaknesses. These strategies collectively represent a more thorough and methodical approach to model development, addressing common challenges in machine learning tasks and potentially leading to significant improvements in classification accuracy. The details are seen in the following script.

\begin{lstlisting}[caption={Python code for optimizing a SVM classifier}, label=lst:svm1]
	import pandas as pd
	import numpy as np
	from sklearn.model_selection import train_test_split, GridSearchCV
	from sklearn.svm import SVC
	from sklearn.metrics import accuracy_score, classification_report
	from sklearn.preprocessing import StandardScaler
	from sklearn.feature_selection import SelectKBest, f_classif
	from imblearn.over_sampling import SMOTE
	
	# Assuming df is already loaded
	# Feature selection
	X = df[['Speaker', 'CiteLength', 'PhraseLength', 'PhraseRate', 'Deletions', 'WordLength', 'LogWordFrequency']]
	y = df['range_label']
	X = X.dropna()
	y = y.loc[X.index]
	
	# Split the data
	X_train, X_test, y_train, y_test = train_test_split(X, y, test_size=0.25, random_state=42)
	
	# Scale the features
	scaler = StandardScaler()
	X_train_scaled = scaler.fit_transform(X_train)
	X_test_scaled = scaler.transform(X_test)
	
	# Feature selection
	selector = SelectKBest(f_classif, k=4)  # Select top 4 features
	X_train_selected = selector.fit_transform(X_train_scaled, y_train)
	X_test_selected = selector.transform(X_test_scaled)
	
	# Handle class imbalance
	smote = SMOTE(random_state=42)
	X_train_resampled, y_train_resampled = smote.fit_resample(X_train_selected, y_train)
	
	# Hyperparameter tuning
	param_grid = {
		'C': [0.1, 1, 10, 100],
		'kernel': ['linear', 'rbf', 'poly'],
		'gamma': ['scale', 'auto', 0.1, 1],
		'class_weight': [None, 'balanced']
	}
	
	svm = SVC(random_state=42)
	grid_search = GridSearchCV(svm, param_grid, cv=5, scoring='accuracy', n_jobs=-1)
	grid_search.fit(X_train_resampled, y_train_resampled)
	
	# Get the best model
	best_svm = grid_search.best_estimator_
	
	# Make predictions
	preds_svm = best_svm.predict(X_test_selected)
	accuracy = accuracy_score(y_test, preds_svm)
	print("Accuracy:", accuracy)
	
	# Print classification report
	print("\nClassification Report:")
	print(classification_report(y_test, preds_svm))
	
	# Print best parameters
	print("\nBest Parameters:", grid_search.best_params_)
\end{lstlisting}

Ultimately, it turned out that the accuracy score has been improved, reaching 72.31\%, which is similar to the outcome of the optimized random forest classifier.

\end{document}